\documentclass[letterpaper, 10 pt, conference]{ieeeconf}  

\IEEEoverridecommandlockouts                              

\usepackage{epsfig}
\usepackage{times}
\usepackage{subfigure}
\usepackage{amsfonts}
\usepackage{amsmath}
\usepackage{amssymb}
\usepackage{graphicx}
\usepackage{url}
\usepackage{psfrag}
\usepackage{array}
\usepackage{mdwmath} 
\usepackage{mdwtab}  
\usepackage{multirow}
\usepackage{verbatim}
\usepackage{color}
\usepackage{textcomp}
\UseRawInputEncoding

\title{\LARGE \bf
Orientation Matters: 6-DoF Autonomous Camera Movement for Minimally Invasive Surgery* 
}

\author{Alaa Eldin Abdelaal$^{1}$ \textit{Student Member, IEEE}, Nancy Hong$^{1}$, Apeksha Avinash$^{1}$, \\ Divya Budihal$^{2}$, Maram Sakr$^{3}$ \textit{Student Member, IEEE}, Gregory D. Hager$^{4}$ \textit{Fellow, IEEE} \\ and Septimiu E. Salcudean$^{1}$ \textit{Fellow, IEEE}
\thanks{This work was supported in part by the Natural Sciences and Engineering Research Council of Canada (Discovery Grant), in part by the Canada Foundation for Innovation (infrastructure and operating funds), in part by Intuitive Surgical (equipment donation), in part by the C.A. Laszlo Chair in Biomedical Engineering held by Prof. Salcudean, and in part by the Vanier Canada Graduate Scholarship held by Alaa Eldin Abdelaal.}
\thanks{$^{1}$A. E. Abdelaal, N. Hong, A. Avinash and S. E. Salcudean are with the Electrical and Computer Engineering Department, University of British Columbia, 2332 Main Mall, Vancouver, BC Canada.
        {\tt\small (email: aabdelaal@ece.ubc.ca)}}%
\thanks{$^{2}$D. Budihal is with the company Zipline International, 333 Corey Way, South San Francisco, CA. Her contribution was made when she was with the Electrical and Computer Engineering Department at University of British Columbia.} 
\thanks{$^{3}$M. Sakr is with the Mechanical Engineering Department, University of British Columbia, 2054-6250 Applied Science Lane,
Vancouver, BC Canada.}       
\thanks{$^{4}$G. D. Hager is with Department of Computer Science, Johns Hopkins University, Baltimore, MD 21218, USA.}
}

\begin{document}

\maketitle
\thispagestyle{empty}
\pagestyle{empty}

\begin{abstract}

We propose a new method for six-degree-of-freedom (6-DoF) autonomous camera movement for minimally invasive surgery, which, unlike previous methods, takes into account both the position and orientation information from structures in the surgical scene. In addition to locating the camera for a good view of the manipulated object, our autonomous camera takes into account workspace constraints, including the horizon and safety constraints. 
We developed a simulation environment to test our method on the ``wire chaser'' surgical training task from validated training curricula in conventional laparoscopy and robot-assisted surgery. Furthermore, we propose, for the first time, the application of the proposed autonomous camera method in video-based surgical skill assessment, an area where videos are typically recorded using fixed cameras. In a study with N=30 human subjects, we show that video examination of the autonomous camera view as it tracks the ring motion over the wire leads to more accurate user error (ring touching the wire) detection than when using a fixed camera view, or camera movement with a fixed orientation.  
Our preliminary work suggests that there are potential benefits to autonomous camera positioning informed by scene orientation, and this can direct designers of automated endoscopes and surgical robotic systems, especially when using chip-on-tip cameras that can be wristed for 6-DoF motion.

\end{abstract}

\section{INTRODUCTION} \label{sec: introduction}

Minimally invasive surgery (MIS) refers to the paradigm of surgery that only needs small incisions into the patient's body to perform complex surgical procedures. Its advantages over open surgery include shorter hospital stay, fewer complications and less pain for the patient. MIS has been successfully used in different surgical specialties such as urology, gynecology and general surgery~\cite{ponsky2012minimally}. Broadly speaking, MIS has two main forms. The first is conventional laparoscopic surgery where 4 degrees-of-freedom (DoF) surgical tools are inserted through small incisions into the body and directly controlled by the surgeon~\cite{bittner2006laparoscopic}. The second is robot-assisted surgery (RAS) where 7-DoF robotic arms with surgical tools are inserted inside the body and controlled by the surgeon from a surgical console of a teleoperation system. The latter comes with many advantages over the former such as easier control of the tools, 3D vision and more precise motion~\cite{abdelaal2020robotics}.    

An important component of MIS platforms is the vision system's endoscopic camera. The camera view is also used by surgeons to infer haptic/force feedback information in RAS~\cite{okamura2011force}. Because camera placement is so important, surgical training curricula usually have a dedicated section on training novice surgeons the skills of moving the endoscopic camera~\cite{stegemann2013fundamental}. Poor handling of the endoscopic camera leads to poor visualization which in turn can disrupt the surgical work flow, prolong the surgical procedures~\cite{mori2015medical}, and compromise the patient's safety~\cite{zhu2013gravity}. Therefore, improvements to the endoscopic camera control system are extremely important as they can improve the overall experience of the surgeon during MIS.  

The current standard practice is manual control of the endoscope motion. In conventional laparoscopic surgery, a dedicated camera assistant is responsible for this task based on the main surgeon's guidance. Such guidance is usually given using verbal communication, making the entire process less efficient, which can lead to suboptimal camera views~\cite{kavoussi1995comparison}. This approach is usually associated with poor ergonomics for the camera assistant who has to hold the endoscope for long periods of time~\cite{lee2009ergonomic}. In robot-assisted surgery, the problems of camera positioning by an assistant are addressed by giving full camera control to the surgeon at the console. This, however, comes with the disadvantage of the surgeon having to switch from controlling the surgical tools to controlling the endoscope, and vice-versa. For surgeries that need many camera motions, such switching of control can be disruptive to the surgeon and can add to his/her cognitive workload. Indeed, with this approach the surgeon needs to control both the tools and the camera, unlike in the conventional laparoscopy case. This can interrupt the surgical work flow~\cite{pandya2014review}.    

To improve the current practice, many groups proposed methods to automate the camera/endoscope motion in MIS. The majority of these methods are based on using some form of tracking information, e.g., tracking the surgeon's tools~\cite{rivas2014towards} or eye gaze~\cite{ali2008eye} or contextual information in the surgical scene itself~\cite{ko2005intelligent}). These camera automation methods only use position information from the structures being tracked, e.g., the position of a specific tool or landmark in the scene.  In this article , we argue that in order to provide a full visual feedback to the surgeon, both position and orientation of structures of interest in the scene should be considered. 

Good visual feedback in MIS does not only improve the performance of the actual surgical tasks, but can also improve video-based surgical skill assessment. Skill assessment is used is to allow surgeons to monitor and evaluate their own (or their trainees') performance in recorded videos with the goal of assessing their skills and identifying areas of improvement in future performances. One aspect of interest is whether the surgeon/trainee mistakenly touches critical areas in the surgical scene. To spot such instances, good visual feedback in the recorded videos is important. We argue that using automated camera methods that consider both the position and orientation of important structures in the surgical scene can improve the accuracy of video-based surgical skill assessment.  

The contributions of this work are as follows:
\begin{itemize}
	\item We propose a novel autonomous camera method that takes the orientation information of the surgical scene into account. In particular, our proposed method is based on following the normal to features of interest in the surgical view. Our proposed method is intended for endoscopic cameras with six DoF. 
	\item We implement the above autonomous camera concept in a simulated environment of the da Vinci surgical system where a pickup camera is attached to one of its arms following our proposed concept in~\cite{avinash2019pickup}. Our implementation includes a motion planning component to satisfy some practical constraints as the camera moves autonomously. 
	\item We propose a novel application of the proposed autonomous camera method in video-based skill assessment in MIS. We compare our method against both a stationary camera and a point-based autonomous camera method where the camera moves to make a point of interest at the center of the view. We evaluate the effectiveness of the proposed method in this application by conducting a user study with N = 30 subjects where subjects assessed the performance of a simulated surgical training task in recorded videos under the above three camera methods. 
\end{itemize}


\section{RELATED WORK} \label{sec: RELATED WORK}

\subsection{Autonomous Cameras in MIS} \label{sub sec: Autonomous Cameras in MIS}

There has been an extensive body of work to facilitate camera control in MIS, with the earlier approaches aiming at giving the surgeon the full control over the camera without the need of any assistance. One of the early projects following this approach was the Automated Endoscopic System for Optimal Positioning (AESOP) project~\cite{allaf1998laparoscopic}. Different control modalities were tested in the context of this project such as voice control~\cite{nathan2006voice} and eye gaze~\cite{ali2008eye}. Head tracking has also been explored in the EndoAssist\textsuperscript{TM} project where the camera moves based on the surgeon's head motion~\cite{gilbert2009endoassist}. While these methods eliminate the need to have a camera assistant, they added to the cognitive load of the surgeon which suggested the need to increase the level of automation in controlling the endoscope.

Automated camera systems in MIS can be categorized based on the source of information used to automate the camera motion into three main categories. The first one is based on the surgical tools. The second is based on the anatomical structures in the surgical scene and the third one is a combination of the first two.  

The majority of the existing work is based on using one or more surgical tools and moving the camera according to their motion. For example, Eslamian~\textit{et al}~\cite{eslamian2020development} propose a method to track two surgical tools and move the camera to make the midpoint of the two tools at the center of the field of view (FOV); they apply their method to the da Vinci surgical robot. A similar approach is proposed in~\cite{casals1996automatic} for conventional laparoscopic surgery. A variation of this method is also proposed in~\cite{rivas2014towards} where the camera moves autonomously to make the currently used tool appear in the FOV (not just the center) based on the current state of the surgical procedure. Moreover, Ma~\textit{et al}~\cite{ma2020visual} track the position of the two tools to control the camera rotation so that the line segment connecting the two tools is always horizontal. Weede~\textit{et al}~\cite{weede2011intelligent} build a Markov model based on the motion of the surgical tools in previous surgeries to predict future motions of these tools. Based on the model, they move the camera's focal point to make the midpoint of the predicted tool positions at the center of the FOV. 

Anatomical structures have also been used to automate the camera motion. In an emulated surgical debridement subtask where the goal is to identify and remove damaged tissue so that the remaining tissue heal faster, Li~\textit{et al}~\cite{ji2018learning} propose a ``learning from demonstration'' approach to automate the camera motion in two dimensions (2D). Their system ranks the damaged tissue and moves the camera's optical axis to make the highest ranked tissue at the center of the FOV.  

The combination of tools and anatomical structures information has also been studied in the context of automating the camera motion in MIS. For example, in~\cite{rivas2017smart}, the authors propose a method that tracks the midpoint of the two tools as well as another anatomical point in the surgical scene. The choice of the additional anatomical point is based on the current state of the surgical task. Ko~\textit{et al}~\cite{ko2005intelligent} build a state transition diagram of the cholecystectomy procedure and propose a method to move the camera based on the current identified state of the procedure. In their method, the camera is moved to make either the currently used tool or a predefined fixed anatomical structure at the center of the FOV. 

In all the above work, the autonomous camera method/algorithm is always based on the {\em position} information of objects of interest in the view. The gap that this work is filling is the use of {\em orientation} information in addition to position to automate the camera motion. In particular, our work explores the use of pose (orientation and position) information of anatomical structures in this context. To the best of our knowledge, this is the first study to explore this aspect in the context of MIS. 

\subsection{Video-based Skill Assessment in MIS}

Video-based methods are extensively used in MIS for training and skill assessment. The effectiveness of these methods has been demonstrated in all surgical settings such as conventional laparoscopy~\cite{singh2015randomized} and RAS~\cite{abdelaal2018play}. Furthermore, these methods have their dedicated and validated skill assessment tools such as the Objective Structured Assessment of Technical Skills (OSATS)~\cite{martin1997objective}.

The effectiveness of video-based skill assessment methods depends on the quality of the visual feedback provided in the videos. One approach to improve the visual feedback is to use multiple cameras that view the surgical scene from multiple perspectives. Several groups have explored the feasibility and effectiveness of this approach as in~\cite{wang2019multiple} and~\cite{abdelaal2020multi}. 

What all the above studies have in common is that the cameras used to record videos are all stationary. In this work, we explore for the first time the effectiveness of using an autonomous camera system based on our proposed method for video-based surgical skill assessment.

\section{PROPOSED METHOD} \label{sec: PROPOSED METHOD}

\subsection{Overview} \label{sub sec: Overview}
In our proposed method, we aim to align the camera such that its optical (or viewing) axis coincides with the normal vector arising from an anatomical structure. Doing so allows the camera to maximize visual coverage of the structure of interest. Additionally, the camera moves to ensure that the anatomical structure is always at the center of the view. In this section, we describe our autonomous camera motion pipeline, various safety measures we have incorporated, and the implementation details of our proposed algorithm. 

\subsection{Motion Pipeline} \label{sub sec: Motion Pipeline}

\begin{figure}
      \centering
     \includegraphics[width=0.4\textwidth]{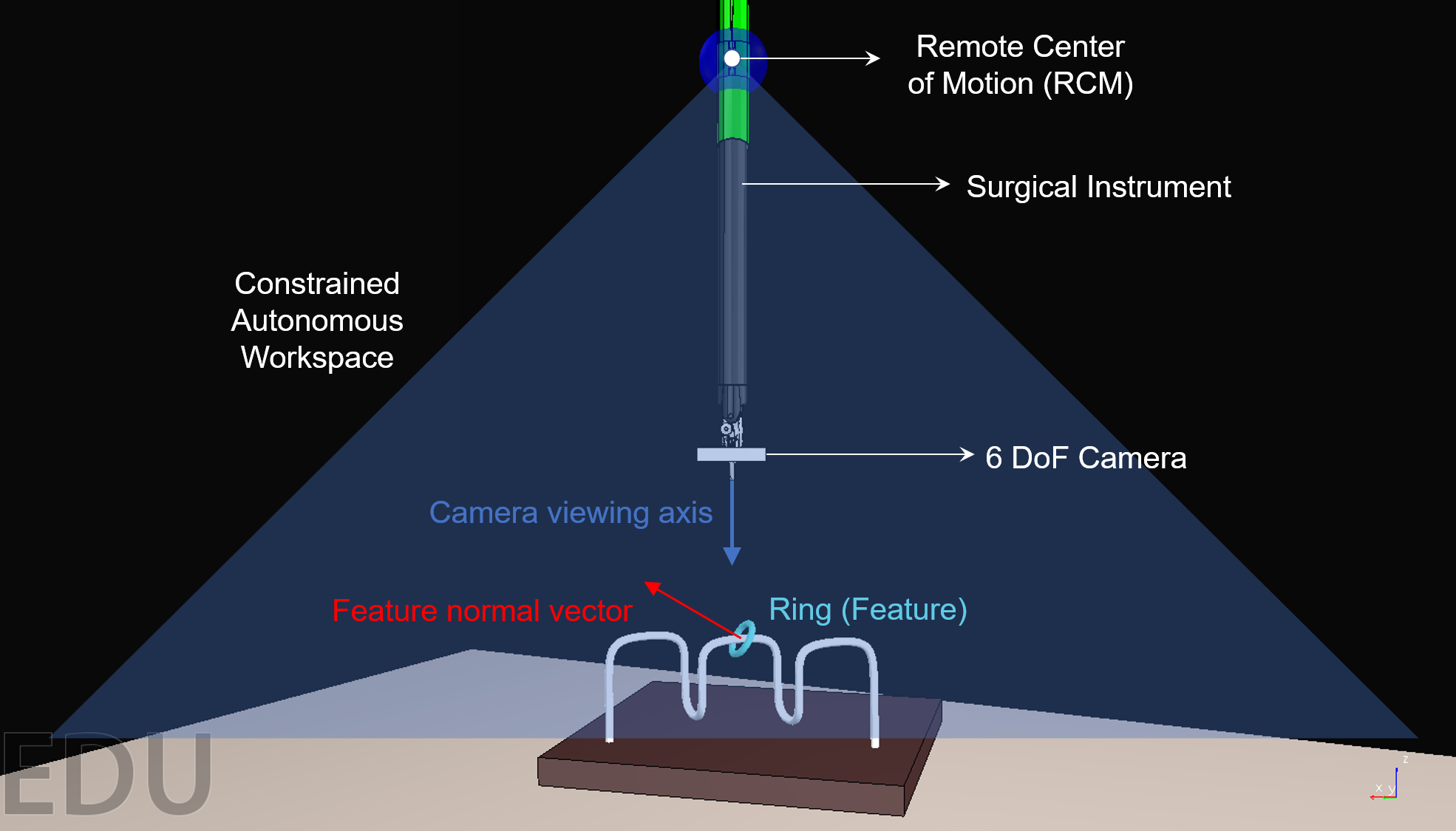}
      \caption{ The setup used with our autonomous camera method showing the wire chaser scene that we added to the simulator. The rail pattern is the same as that in the validated curriculum in~\cite{schreuder2011laparoscopic}.}
      \label{constrained_auto_workspace}
   \end{figure}

We apply our motion pipeline to the setup shown in Fig.~\ref{constrained_auto_workspace}. To center the anatomical structure of interest in the camera view, we consider the feature's position $\textit{\textbf{p}}_c$, and its normal vector $\textbf{n}$. We compute a goal position $\textit{\textbf{p}}_g$ along this normal vector, at a fixed distance $\textit{d}_f$ from $\textit{\textbf{p}}_c$. To avoid collision with tissue, we consider only the space above the anatomical feature and accordingly consider either $\textbf{n}$ or $-\textbf{n}$. At each instant, the camera's positional goal is set to be the computed point $\textbf{\textit{p}}_g$ such that $\textit{\textbf{p}}_g = \textbf{\textit{p}}_c \pm \textit{d}_f \textbf{n}$.

\begin{figure}
      \centering
     \includegraphics[width=0.2\textwidth]{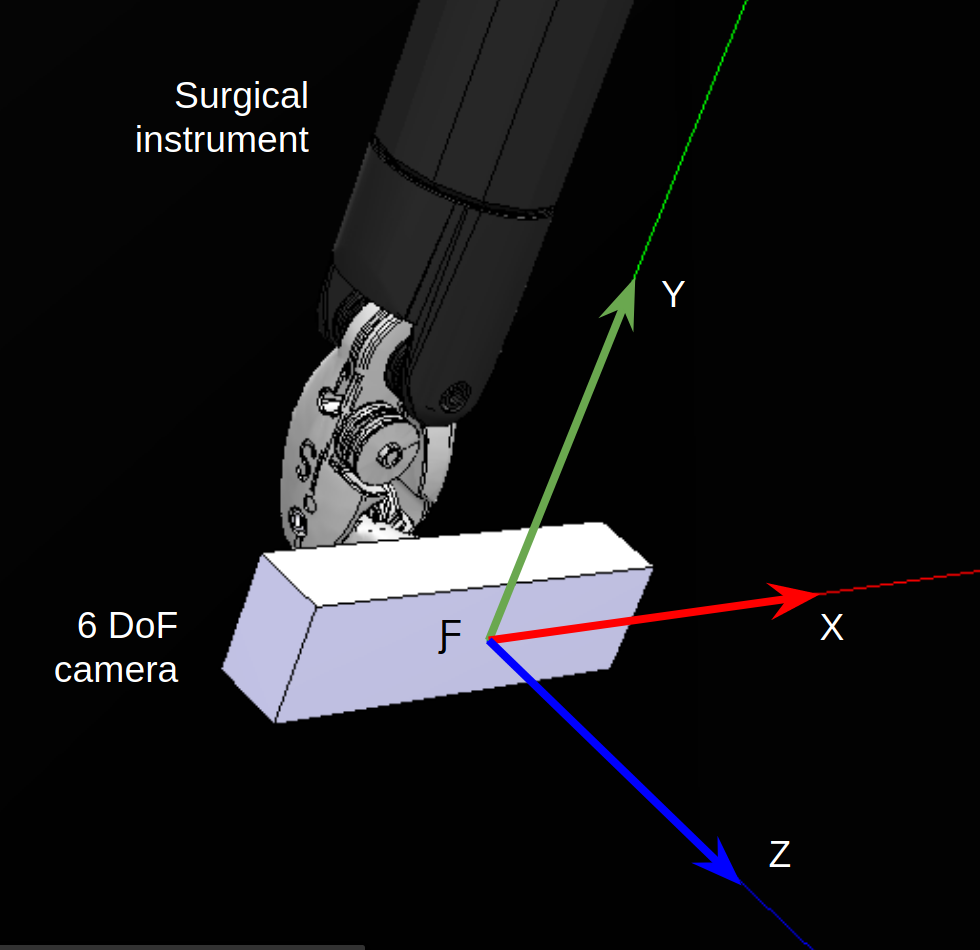}
      \caption{Frame $\digamma$ attached to the camera.}
      \label{camera_frame_labeled}
   \end{figure}

The camera's orientation can be fully described by a frame $\digamma$ attached to it, as seen in Fig.~\ref{camera_frame_labeled}, and mathematically represented by a rotation matrix $\textbf{R}_g$ where $col_1$[$\textbf{R}_g$] represents the \textit{x-axis}, $col_2$[$\textbf{R}_g$] the \textit{y-axis}, and $col_3$[$\textbf{R}_g$] the\textit{ z-axis} attached to the camera. The primary goal of our algorithm is to align the camera's optical axis with $\textbf{n}$, achieved by setting the \textit{z-axis} of frame $\digamma$ to $\textbf{n}$. Another desirable characteristic when moving the camera in surgery is to maintain a correct camera horizon~\cite{shetty2012construct}, and this is controlled by the camera's horizontal (or \textit{x}) axis. We set this \textit{x-axis} to be the cross product between a vector that is pointing upwards (i.e. \textit{z-axis} of the world frame) and the feature normal vector $\textbf{n}$ so that we always obtain an \textit{x-axis} that is parallel to the xy-plane of the world frame. The \textit{y-axis} is simply chosen to be orthonormal to the other two axes and this completes the desired rotation matrix $\textbf{R}_g$.

\begin{figure}
      \centering
     \includegraphics[width=0.4\textwidth]{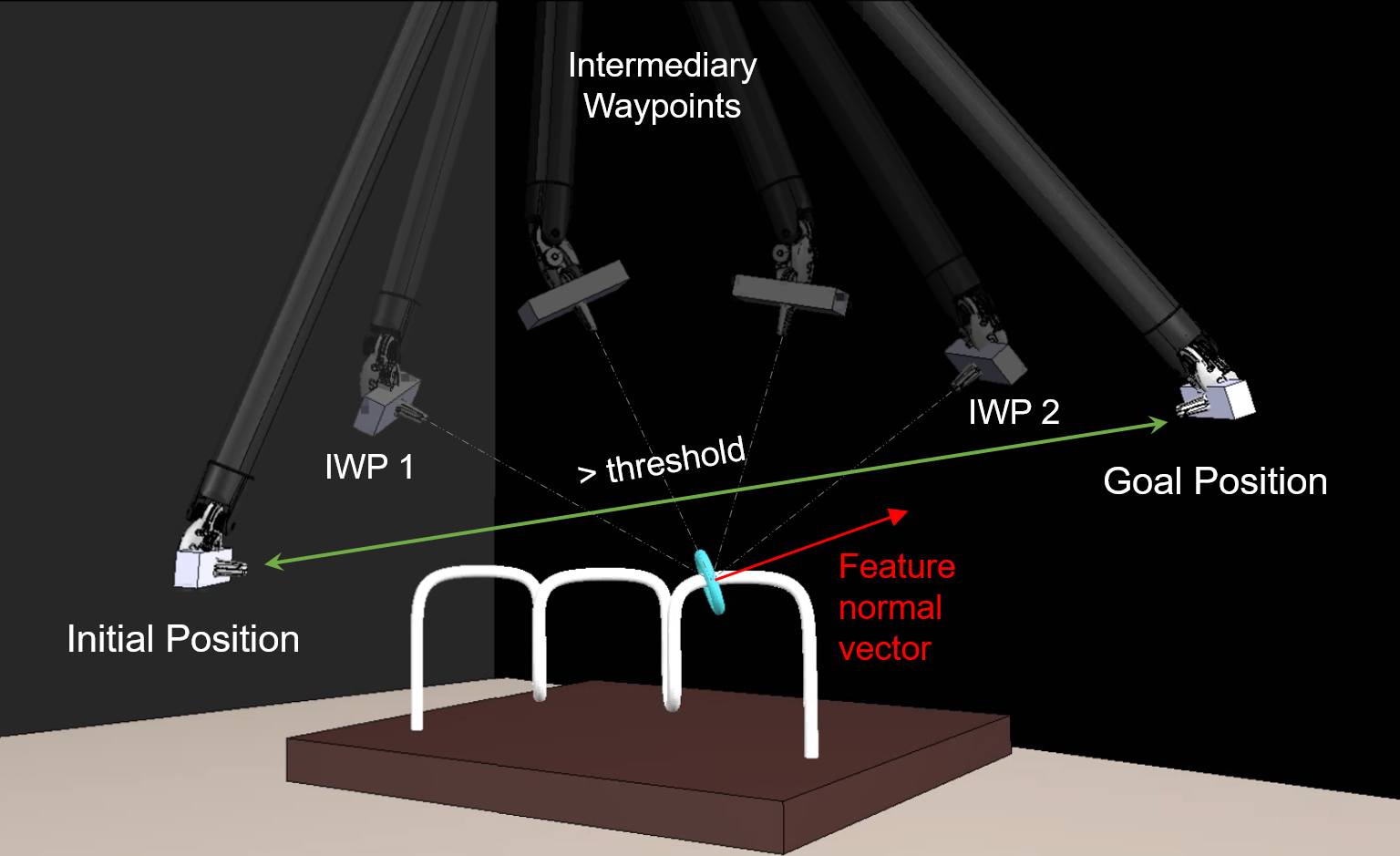}
      \caption{The motion planning part of our autonomous camera method where IWP are generated.}
      \label{motion_planning}
   \end{figure}

To facilitate collision avoidance, motion planning is incorporated when the distance between consecutive goal positions is larger than a set threshold.  We generate a set of intermediate waypoints (IWP) between the current camera position and the goal position as seen in Fig.~\ref{motion_planning}. IWP 1 and 2 are manually set to a fixed distance $\textit{d}_{wp}$ above the current and goal positions, respectively, along the positive \textit{$z$-axis} of the world frame (or any vector pointing upwards). Using these four points, a linear function is interpolated and a trajectory is obtained. The orientation of the camera at each of these new intermediate positions is adjusted to ensure that the feature is always centered in the view. The \textit{$z$-axis} is set to the vector between the current intermediate point and the feature's center point. The \textit{$x$-axis} is adjusted to correct for the horizon as described previously, and the \textit{$y$-axis} is chosen to be a vector orthonormal to both. 

\subsection{Safety Measures} \label{sub sec: Safety Measures}

Without human-in-the-loop control, it is essential to incorporate safety features into any autonomous system. First and foremost, we define a constrained workspace ((see Fig.\ref{constrained_auto_workspace})) within which we autonomously control the camera. Outside this workspace, the autonomous algorithm freezes, allowing the surgeon to manually control the camera according to his/her discretion. We chose to define this constrained workspace in the form of a 3D cone, with the following parameters: the cone tip is the remote center of motion (RCM), the cone height is slightly smaller than the length of the surgical instrument/endoscope, and the cone base radius is empirically chosen to be 10 cm. The cone's directional vector is initially set to the vector joining the RCM and the initial position of the feature, to ensure that the feature of interest is always in view, and remains unchanged after initialization.

An additional safety constraint is to ensure that all surgical tools are always in the FOV. An out-of-view tool can unknowingly bring severe and unwanted damage to tissue in the surrounding area. Given the instrument tip positions in 3D space and using the camera's intrinsic parameters, these tip positions can be computed in image space at each frame. We define two windows within the image: an outer window, and an inner window,  and adjust the camera's distance to the structure of interest until all surgical tools are found to be within the space between these windows. The adjustment is made by incrementing the distance $\textit{d}_f$ in our autonomous camera algorithm by $\pm 1$ mm accordingly.

\subsection{Implementation Details} \label{sub sec: Implementation Details}

Our proposed algorithm is implemented on a simulated da Vinci surgical system, where visual feedback is provided with a stereo endoscope. The 4-DoF endoscope provides limited angular or orientational freedom of movement, and hence cannot be fully exploited to show the merits of our proposed algorithm. We instead present our implementation with a 6-DoF stereo camera that is attached to the end of the surgical tool tip. Another possibility is to use the ``pickup'' stereoscopic camera concept proposed in our previous work~\cite{avinash2019pickup}. The pickup camera is inserted axially through a surgical incision into the patient's body and can be picked up and controlled by a surgical instrument (such as the da Vinci ProGrasp forceps) through its grasping interface.  
	
For this work, we focus on demonstrating the advantages of our proposed algorithm, and hence obtain and use ground truth data such as position and normal vector of the anatomical structure of interest. For future implementations, this data can be obtained through a dedicated vision pipeline as in~\cite{yip2012tissue}. It should be noted that any appropriate inverse kinematics module can easily be used to implement the proposed algorithm on any other camera-based robotic system such as articulated and snake-like cameras~\cite{yeung2012technical}. 
	
\section{EXPERIMENTAL EVALUATION} \label{sec: EXPERIMENTAL EVALUATION}

\subsection{Experimental setup} 

Due to the COVID-19 situation, we tested our proposed method in a simulated environment, instead of using the da Vinci Research Kit (dVRK)~\cite{kazanzides2014open}, as originally planned. We use the first generation da Vinci system simulator proposed in~\cite{fontanelli2018v}. It simulates the full patient-side cart of the da Vinci system including two patient-side manipulators (PSMs) and a 4-degree-of-freedom endoscopic camera manipulator (ECM). Controlling the motions of the PSMs and ECM can be performed in the same way as controlling the patient-side cart in the real robot using the dVRK. The simulator also includes an interface with the Robot Operating System (ROS). The simulator comes with some pre-prepared scenes of different tasks and it also allows adding new scenes/environments as needed. We used this feature to add the wire chaser task scene as shown in Fig.~\ref{constrained_auto_workspace}, which is described in~\ref{sub sec: Task}. We modified the simulated da Vinci system as described in~\ref{sub sec: Implementation Details} to include a 6-DoF endoscope.

\subsection{Task} \label{sub sec: Task}

We test our autonomous camera method on the ``wire chaser'' task which is part of the validated training curricula in conventional laparoscopy~\cite{schreuder2011laparoscopic} as well as RAS~\cite{stegemann2013fundamental}. The task has also been validated for multiple surgical specialties such as urology, gynecology and general surgery~\cite{alzahrani2013validation}. Previous research has shown that the level of performance in this task is correlated with the performance level in the operating room~\cite{aghazadeh2016performance}. 

The task involves holding a ring and moving it along a rail/wire. It is designed to measure the manual dexterity, hand-eye coordination and camera control skills of trainees. In our version of this task, trainees are penalized if the ring touches the rail. The same version of the task has also been used in the context of robot-assisted surgical training as in~\cite{mariani2020experimental}.

The ring represents the anatomical structure that we are interested in. The main idea is that a good visualization of the ring (as seen from the camera) is crucial to the wire chaser task. That is why our autonomous camera method used both the plane of the face of the ring as well as the ring's center as its inputs. The camera then moves autonomously following our proposed method in Section~\ref{sec: PROPOSED METHOD} so that the viewing plane of the camera is always parallel to the plane of the face of the ring and that the camera focal point is always at the center of the ring. 

Our hypothesis is that using the proposed autonomous camera method, subjects can better spot the cases when the ring touches the rail compared with other methods that automate the camera based only on position information. The position information in this case is the position of the center of the ring. This hypothesis is tested in the context of video-based skill assessment where subjects watch videos of the task and their goal is to assess the skill using specific criteria.  

Towards this end, we recorded several videos of the wire chaser task. We automated the ring motion that is held by one PSM to follow predefined trajectories along the rail. Some of these trajectories are ideal according to the following two conditions: (i) The ring is centered with respect to the rail and (ii) The ring's face plane is always perpendicular to the rail. Other trajectories were randomized by violating one or two of the above conditions. This in turn introduces a number of collisions between the ring and rail. 

The trajectory of the rail is defined by setting control points evenly spaced from start to finish. A control point's position is defined in ($x$, $y$, $z$) coordinates, and orientation in Tait-Bryan Euler angles ($\alpha, \beta, \gamma$) that together represent a single rotation: $\textbf{R}_{total}$ = $\textbf{R}_{x}(\alpha)$ $\textbf{R}_y(\beta)$ $\textbf{R}_z(\gamma)$. $\textbf{R}_z$, $\textbf{R}_y$ and $\textbf{R}_x$ represent elemental rotations about the \textit{z}-, \textit{y}- and \textit{x}-axes respectively of the simulation world frame. We automate the ring movement in the simulator to follow the pose of these control points. To introduce collisions between the ring and rail, randomly generated noise is added to the six variables describing each of the control points. By varying the number of control points and the threshold of noise added to the position and orientation, varying levels of difficulty can be represented in the resultant trajectories. For our tests, we chose trajectories with the following parameters: control points: 36 and 71, position threshold: 3-3.5 mm, angular threshold: 10-30 degrees. A higher number of control points introduces a higher degree of variation in the trajectory, providing more touches/collisions between the ring and rail resulting in the trajectory with 71 control points being the most difficult one. 

We tested the proposed autonomous camera method in two aspects. The first one is by measuring the tracking errors  as the ring moves along the ideal and randomized trajectories which are explained in~\ref{subsec: Self-reported Metrics} below. The second one is by conducting a user study where users watch the recorded videos to count the number of touches between the ring and rail. The goal in this second case is to measure how accurate the users are when using the proposed method in comparison with other methods as explained in~\ref{subsec: User Study} below. 

\subsection{Tracking Performance Metrics} \label{subsec: Self-reported Metrics}

The first method to evaluate the proposed autonomous camera method is to measure the tracking accuracy when the ring is moving along ideal and randomized trajectories. There were two sources of errors in our setup. The first one is the lag between the ring motion and camera motion, as we did not incorporate any information about future ring path into our autonomous camera method. The second is the limited performance of the simulator on our computing platform. 
We consider the following three tracking errors: 
\begin{itemize}
	\item Centering error in the image space: This metric refers to the difference in pixels between the position of the center of the ring on the camera view and the position of the center of the view.
	\item Centering error in the 3D space: This metric is similar to the first one except that it is the difference in millimeters between the 3D position of the center of the ring and the equivalent 3D position of the center of the FOV.
	\item Orientation error: This metric refers to the angle between the camera optical axis and the vector $\textbf{n}$ that is perpendicular to the plane of the face of the ring.
\end{itemize}

 The above three errors are reported as a function of time along the entire trajectory of the ring. We report them in three cases of using ideal trajectories and in another three cases of using randomized ones. 
 
\subsection{User Study} \label{subsec: User Study}

We conducted a user study (N = 30) to measure the effectiveness of the proposed autonomous camera method while performing a video-based skill assessment task as described above. We recorded videos of the wire chaser task as the ring follows different randomized trajectories under three conditions for the camera motion as follows: 
\begin{itemize}
	\item Condition I is when the camera is fixed, showing the entire task. This represents the baseline condition. 
	\item Condition II is when the camera motion is automated to follow the center point of the ring, regardless of the ring's orientation. This represents an autonomous camera method that is based solely on position information. We refer to this method as the ``centering method''.  
	\item Condition III is when the proposed autonomous camera method is applied. That is, when the autonomous camera method is based on both the position and orientation information as described in Section~\ref{sec: PROPOSED METHOD} above.  
\end{itemize}
In the last two conditions, the camera initial pose was the same as in the fixed camera condition (condition I) above.

This was a within-subject user study, where each subject was exposed to all the study conditions. We recorded a total of nine videos, three per each condition. The nine videos were for the ring moving along the rail in three randomized trajectories with varying levels of difficulty. The goal is to measure the skill assessment accuracy in the videos where the ring moves in the most difficult trajectory (that is, the one with highest level of randomness which is the trajectory that has 71 control points). Each subject watched the nine videos in three sets, each set containing the three videos of each condition. Subjects were asked to count the number of touches between the ring and rail in each video. Counterbalancing was employed to reduce/eliminate the effect of any learning or carryover bias that may exist when a subject is exposed to each condition. The Latin squares~\cite{mackenzie2002within} method was used to compute the order in which each subject is exposed to a condition. Since the study has three conditions, we applied two Latin squares, the second one being the mirror of the first, which led to having a total of six cases representing all the six possible combinations of the three conditions.

Due to the restrictions of inviting subjects to the lab (because of the COVID-19 situation), the study was conducted virtually by sending an electronic form to each subject containing the videos. We added an attention question in the middle of the form to make sure that subjects were paying attention. Any subject who provided a wrong answer to this question was excluded from the study and his/her data were not considered. 

All our subjects were university students with little or no exposure to surgery. Previous research shows that crowd-sourced video-based surgical skill assessment with non-experts (as we did in this user study) is as accurate as the skill assessment performed by expert surgeons and surgical educators~\cite{chen2014crowd}. The user study was approved by the Research Ethics Board at the University of British Columbia.

\section{RESULTS} \label{sec: RESULTS}

\begin{figure}
      \centering
     \includegraphics[width=0.4\textwidth]{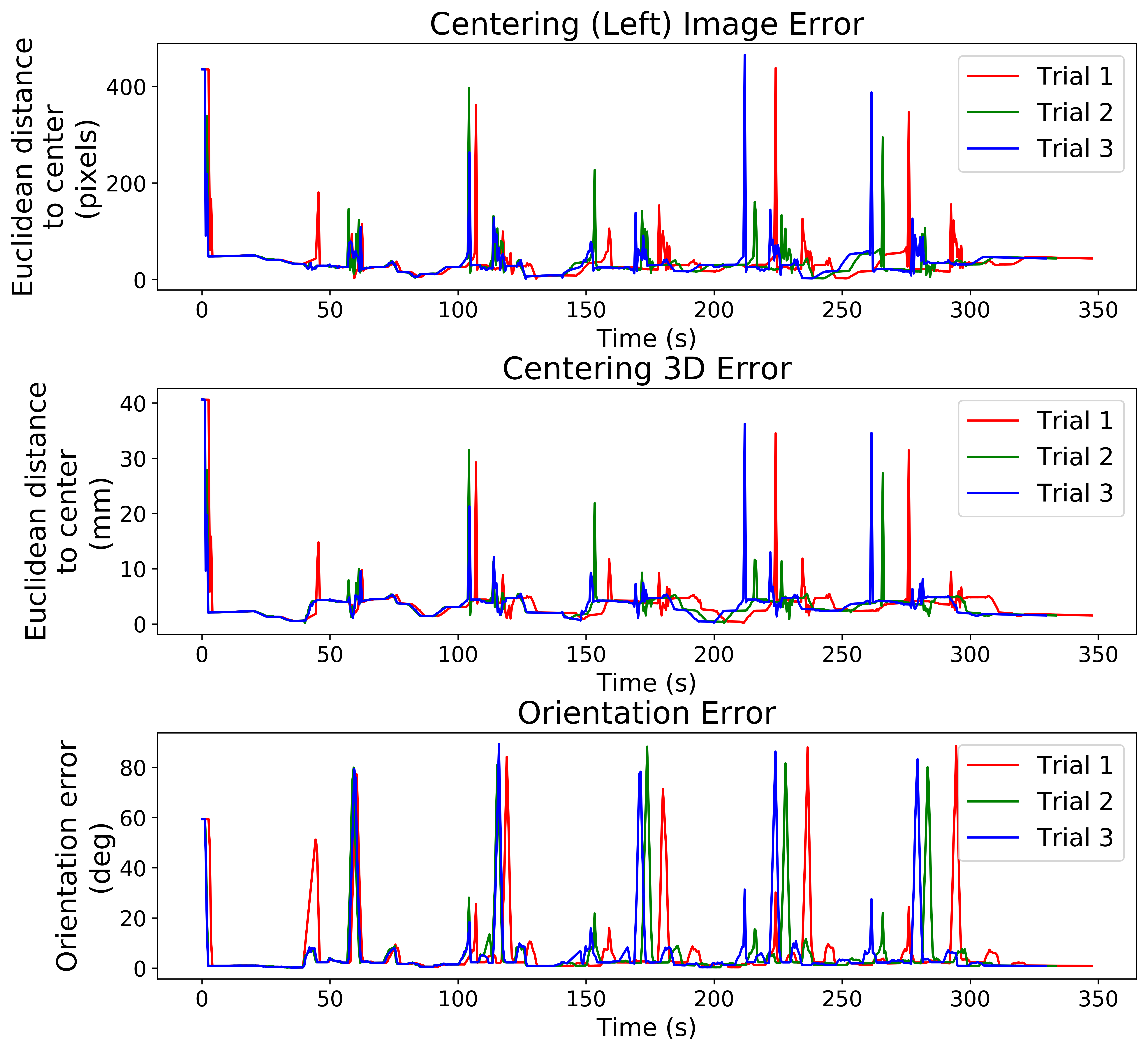}
      \caption{The tracking accuracy results of the proposed autonomous camera method in three ideal trajectories.}
      \label{tracking_path_original}
   \end{figure}

Based on the performance metrics outlined in~\ref{subsec: Self-reported Metrics}, we conducted two tests with the wire chaser task to evaluate the accuracy of our implemented algorithm. In the first test, the trajectory represents the ideal trajectory of the ring along the rail, without any collisions/touches between the two. Our 6-DoF camera follows the ring at a fixed distance, with its optical axis aligned with the ring's normal vector. We compute the three metrics: centering error with respect to the left image space, centering error in the 3D space, and orientation error, as shown in Fig.~\ref{tracking_path_original}. This test is repeated three times (represented by trials 1, 2, and 3) to show the repeatability of our algorithm. Across all three trials, we obtained an average image centering error of 35 pixels, 3D centering error of 3.41 mm, and orientation error of 5.45 degrees. The peaks noticeable in the plots correspond to segments of the trajectory where we implement motion planning (due to largely distanced consecutive goal positions); to ensure that the feature is always in view. In these cases, we relax our orientation constraint which leads to large reported errors in the orientation angle. Despite these peaks, the average overall tracking accuracy of our system remains high.  

\begin{figure}
      \centering
     \includegraphics[width=0.4\textwidth]{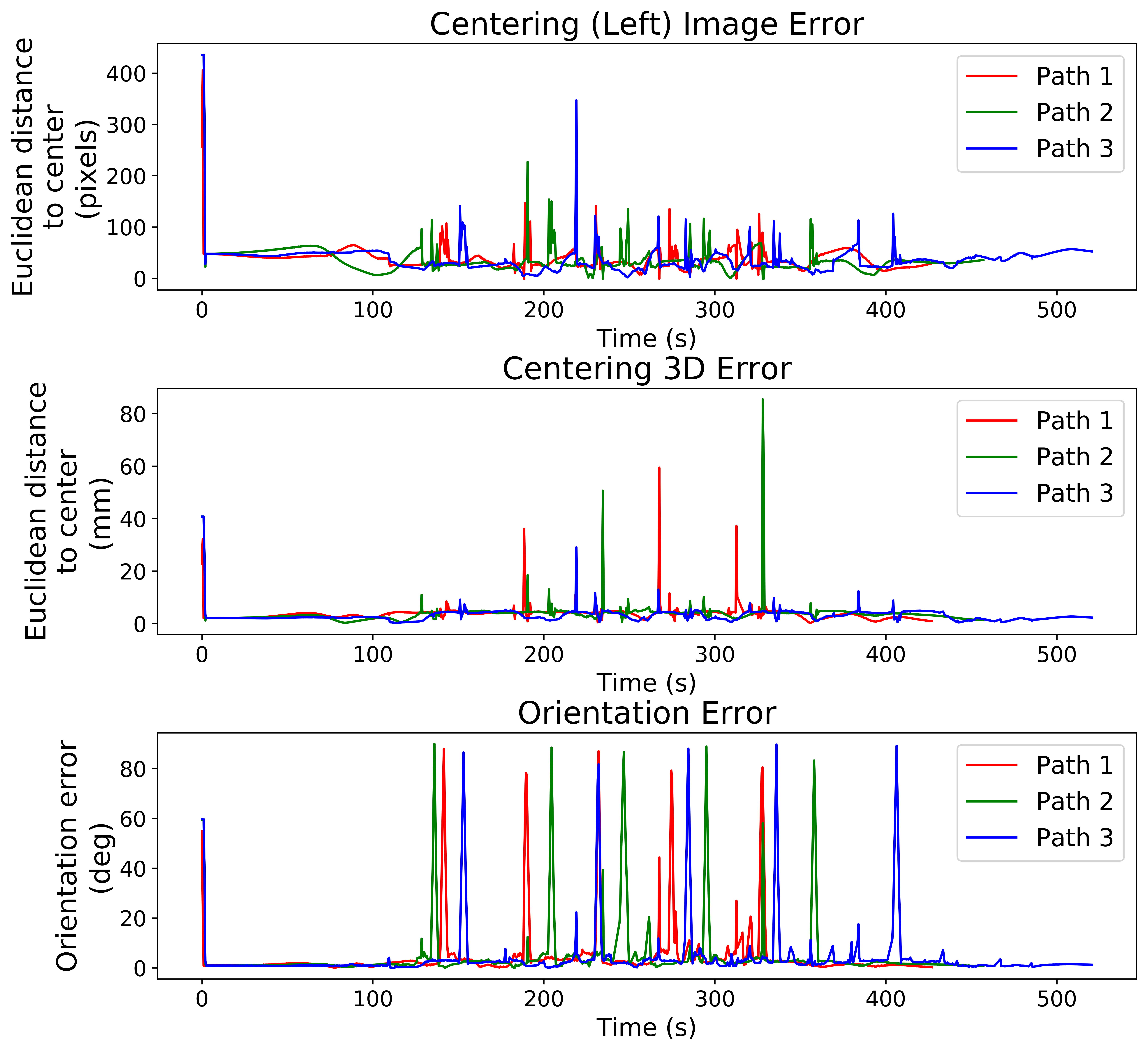}
      \caption{The tracking accuracy results of the proposed autonomous camera method in three randomized trajectories.}
      \label{tracking_paths_noisy}
   \end{figure}

	In the second test, we chose three noisy trajectories by adding noise to the ideal positions and orientations of the ring across its path such that the ring collides with the rail at certain points. The three noisy trajectories are the same trajectories used in the user study described in~\ref{subsec: User Study}, and are represented by Path 1, 2 and 3 in Fig.~\ref{tracking_paths_noisy}. We obtained an average image centering error of 36 pixels, 3D centering error of 3.33 mm, and orientation error of 4.30 degrees across the three paths. Similar to Fig.~\ref{tracking_path_original}, the peaks shown in Fig.~\ref{tracking_paths_noisy} correspond to the motion planning segment with relaxed orientation constraints; the peaks here occur at different points of time for the three paths and hence appear more spread out.
	
	For both these tests, the errors across all three performance metrics are very low, indicating the high accuracy of our implementation, as can be seen in Figs.~\ref{tracking_path_original} and~\ref{tracking_paths_noisy}. It must be noted that the average error reported above includes the errors from the peaks, but even so, the final errors are reasonably low.

As for the user study, we report the assessment errors in the most difficult video, that is, the one with the highest level of randomization. Assessment errors refer to the absolute difference between ground truth errors (which we get from the simulator) and the reported errors by each subject. From the 30 participants in the user study, three participants were excluded after providing a wrong answer to the attention question. In the remaining data, outliers have also been identified and removed. We then compared between the subjects' assessment errors across the three study conditions. 

\begin{figure}
      \centering
     \includegraphics[width=0.4\textwidth]{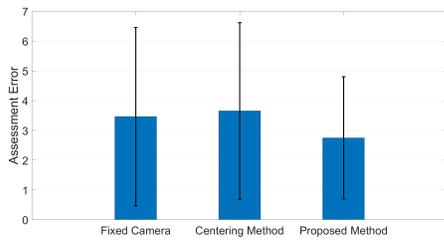}
      \caption{The results of the user study based on the assessment errors of the subjects in the most difficult skill assessment video.}
      \label{assessment_error_video3}
   \end{figure}

As shown in Fig.~\ref{assessment_error_video3}, using the proposed autonomous camera method leads to lower number of assessment errors and less variance between the subjects' scores compared with the other two conditions. In particular, the proposed method (condition III) leads to 25\% and 21\% fewer assessment errors compared with the centering method (condition II) and fixed camera method (condition I), respectively. Furthermore, the standard deviation in the assessment errors using the proposed method is 31\% and 32\% lower than that of the centering method and fixed camera method, respectively. 

 These reductions in the average and standard deviation of the proposed method show its potential to improve the current practice in video-based surgical skill assessment. Previous research in this area show that variability between assessors is a major practical problem. Gingerich~\textit{et al}~\cite{gingerich2014seeing} report that this variability comes from the cognitive limitations of the assessors as well as from their making of unjustified inferences. Our proposed autonomous camera method has the potential to contribute to solving these problems as it can provide better visual feedback which allows the assessors to make more informed assessments and reduces their need to infer/guess due to the lack of the available visual information. 

\section{DISCUSSION} \label{sec: DISCUSSION}

Orientation matters in viewing the surgical scene. Instructions of many surgical procedures include moving the camera to view specific anatomical structures in a predefined orientation. Our proposed method provides an automated way of achieving this, which can reduce the burden of controlling the camera from the surgeon. This is especially true because adjusting the view with respect to orientation requirements is arguably more difficult to achieve by manually controlling the camera than adjusting with respect to the position only. This is a more difficult problem when controlling articulated/snake-like endoscopes~\cite{ma2020visual} and we believe that our proposed method is a first step towards solving this problem.

We view our proposed autonomous camera method in the bigger context as a component of an intelligent assistant to the surgeon. Such an assistant can use the surgical data to recognize the current stage/step of the surgical work flow. It can then use this to infer the anatomical structures of interest automatically and with input from the surgeon if needed. It can also provide the visualization requirements of these anatomical structures in terms of their required position and orientation. Computer vision methods can be used to identify and recognize these structures in the scene. Our proposed method can then be used to realize the required visualization requirements. The resulting camera views can then be transferred back to the intelligent assistant which can use them to infer the new step/stage of the surgical work flow.  

Each of the above components of the intelligent assistant is an active research area on its own (e.g., surgical work flow analysis~\cite{stauder2014random} and tissue identification~\cite{yip2012tissue}). Our proposed system provides another component which differs from previous work in the flexibility it provides in satisfying a wider range of visualization requirements.

\section{CONCLUSIONS} \label{sec: CONCLUSIONS}

We presented an autonomous camera method for 6-DoF endoscopic camera systems in MIS. Our method takes into consideration both the position and orientation information of anatomical structures of interest in the surgical scene. Our method achieved an average position tracking accuracy of 3 mm and orientation tracking accuracy of 5 degrees when tested on a validated MIS training task in a simulated environment. We also presented some safety measures that can be included into our autonomous camera system to avoid collisions with anatomical structures in the surgical scene and avoid having the surgical tools outside the FOV.   

We also tested the effectiveness of using an autonomous camera system for video-based surgical skill assessment. We conducted a user study (N = 30) where subjects watched videos of a simulated surgical training task under different camera motion/automation conditions. Our results show that using the proposed autonomous camera method leads to up to 25\% more accurate skill assessment and up to 32\% lower standard deviations between different assessors. These results demonstrate the potential of the proposed autonomous camera method in augmenting the cognitive abilities of assessors by providing better visual feedback of the tasks compared with the other methods.     

Our results show the importance of including orientation information into automated camera system in MIS. With the extensive research on articulated endoscopic cameras, the constraints of tracking such information in practice are removed, unlike the commonly used 4-DoF endoscopic systems. Our future work includes improving the proposed autonomous camera pipeline to consider more than one anatomical structure, addressing potential problems in the visual feedback such as occlusions, and testing the proposed system with subjects conducting a surgical task on a MIS platform such as the da Vinci system.


\section*{ACKNOWLEDGMENT}

We would like to thank Jordan Liu for his assistance with use of the simulator for this work.

\bibliographystyle{IEEEtran}
\bibliography{./refs}    

\end{document}